\pdfoutput=1

\documentclass[conference, compsoc]{IEEEtran}

\usepackage{graphics}
\usepackage{epsfig} 
\usepackage{subfigure}

\begin{document}

\title{PyraPose: Feature Pyramids for Fast and Accurate Object Pose Estimation under Domain Shift}

\author{
\IEEEauthorblockN{Stefan Thalhammer, Markus Leitner, Timothy Patten and Markus Vincze}
\IEEEauthorblockA{Faculty of Electrical Engineering and Information Technology\\
TU Wien, 1040 Vienna, Austria \\
email: {thalhammer, leitner, patten, vincze\}@acin.tuwien.ac.at}}
}

\maketitle

\begin{abstract}

Object pose estimation enables robots to understand and interact with their environments.
Training with synthetic data is necessary in order to adapt to novel situations. Unfortunately, pose estimation under domain shift, i.e., training on synthetic data and testing in the real world, is challenging.
Deep learning-based approaches currently perform best when using encoder-decoder networks but typically do not generalize to new scenarios with different scene characteristics.
We argue that patch-based approaches, instead of encoder-decoder networks, are more suited for synthetic-to-real transfer because local to global object information is better represented.
To that end, we present a novel approach based on a specialized feature pyramid network to compute multi-scale features for creating pose hypotheses on different feature map resolutions in parallel.
Our single-shot pose estimation approach is evaluated on multiple standard datasets and outperforms the state of the art by up to $\sim$35~\%. We also perform grasping experiments in the real world to demonstrate the advantage of using synthetic data to generalize to novel environments.



\end{abstract}

\section{Introduction}



Object pose estimation is important for many robotic tasks that require scene understanding and object manipulation. 
However, in practical applications the places of deployment are inherently diverse and often arbitrary. This is detrimental for CNN-based methods trained on real-world data because of the change in background, illumination, object appearance and other scene characteristics, which results in reduced performance in novel domains~\cite{kaskman2019homebreweddb}. An additional challenge is changing sets of objects meaning that re-training the pose estimator becomes necessary. However, capturing real-world training data and manually annotating these every time the setup changes is infeasible.


Thus, exploiting synthetic data is a popular direction in order to generalize to novel domains and to fully automate the training procedure. 
The best performing deep-learning approaches for single-shot object pose estimation employ encoder-decoder architectures~\cite{hu2019segmentation, li2019CDPN, park2019pix2pose, peng2019pvnet, zakharov2019dpod}. 
These methods only focus on local changes in image space, which is detrimental for pose estimation under domain shift, i.e., synthetic and real. 
Alternatively, coalescing features from multiple spatial resolutions~\cite{liu2016ssd} enables processing of local to global information in parallel. This is more representative and therefore can lead to better generalization as demonstrated for object detection, where multi-scale features are computed using feature pyramids~\cite{ghiasi2019fpn, lin2017feature, linRetina, liu2018path, tan2019efficientdet}.

\begin{figure}
   \centering
   \includegraphics[width=0.97\columnwidth]{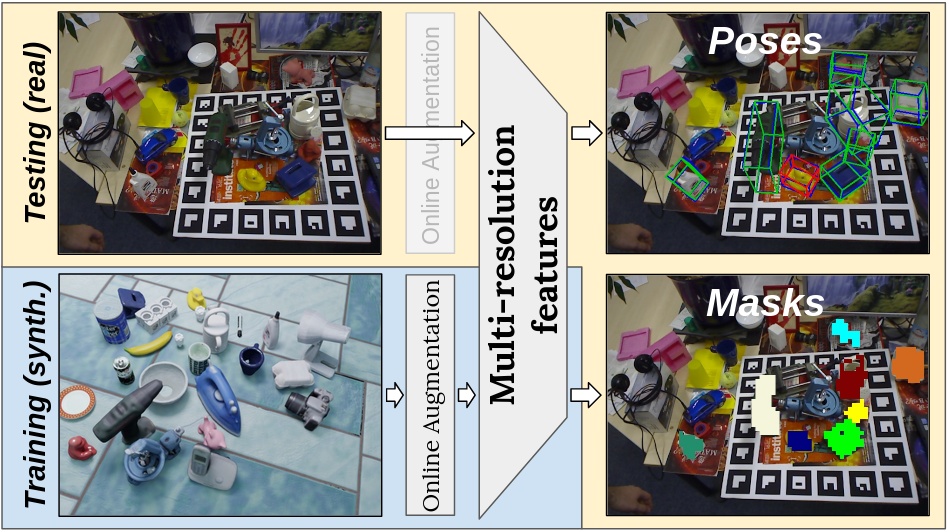}
   \caption{Single-shot multi-object pose estimation. Trained on augmented synthetic data. At test time, predictions on real-world data to estimate 6D poses and segmentation masks.}
   \label{fig:ovview}
   \vspace{-2.0ex}
\end{figure}

This paper proposes using feature pyramids as a main building block for extracting meaningful hierarchical features for pose estimation in contrast to those generated by encoder-decoder networks. As such, we show that making predictions from multi-resolution feature maps guides the network to learn robust pose estimation under occlusion and domain shift. 
Our approach PyraPose, shown in Figure~\ref{fig:ovview}, is based on a novel Pose Feature Pyramid Network (PFPN), specialized for pose estimation.
Multi-resolution features are created and task heads estimate instance segmentation and 2D-3D correspondences to estimate object poses with the Perspective-n-Points (PnP) algorithm. Our design needs only one network (less than 43M parameters) per set of objects, which leads to a fast inference time of $\sim$26 fps. In comparison to state-of-the-art methods trained on synthetic data, we achieve up to $\sim$35\% higher accuracy. Furthermore, real-world grasping experiments with a mobile manipulator demonstrate the full capability of the synthetic-to-real transfer.

In summary, our contributions are the following:
\begin{itemize}
\item A novel approach for pose estimation based on a new feature pyramid network, PFPN, that creates meaningful multi-scale features for pose estimation.
\item The demonstration that multi-scale features are well suited for pose estimation under domain shift, allowing generalization to different domains of deployment by only training on synthetic data.
\item State-of-the-art performance on multiple standard datasets and demonstration of the usability for real-world robot grasping.
\end{itemize}

The remainder of the paper is structured as follows. Section~\ref{sec:related_work} discusses related work. This is followed by a description of our proposed approach in Section~\ref{sec:dmpose} and evaluations in Section~\ref{sec:experiments}. Lastly, Section~\ref{sec:conclusion} concludes the paper with a summary and discussion.

\section{Related Work}\label{sec:related_work}

This section presents the related work on single-shot object pose estimation, hierarchical feature aggregation and dealing with domain shift for pose estimation.

\subsection{Pose Estimation}

A large variety of approaches are applied for CNN-based pose estimation.
The dominant strategy is to employ encoder-decoder architectures for dense hypotheses generation and subsequent pose estimation using the PnP algorithm~\cite{he2019pvn3d, Hodan2020EPOSE6, li2019CDPN, oberweger2018making, park2019pix2pose, peng2019pvnet, wang2019densefusion, zakharov2019dpod}.
The majority of these approaches color the mesh model using \textit{uv}-coordinates.
The networks are trained to regress the vertex location in the object coordinate system, i.e., 2D-3D correspondences, that are the input to PnP.
Alternatively to this, a family of approaches exists that is more closely related to object detection~\cite{hu2019segmentation, oberweger2018making, rad2017bb8, tekin2018real}. 
These methods either predict a single or multiple 3D bounding box hypotheses~\cite{hu2019segmentation, oberweger2018making, rad2017bb8, tekin2018real} as 2D-3D correspondences.
Recent patch-based approaches such as~\cite{hu2019segmentation} and~\cite{tekin2018real} have fast inference time, however, have inferior performance to the encoder-decoder approaches. 
This performance gap occurs because these approaches do not coalesce features of different spatial resolutions and thus miss object scale information.


We therefore address this shortcoming by making predictions from multi-scale features of image patches, which yields strong results under domain shift and improved performance over encoder-decoder approaches.
We a) aggregate multi-scale features using a feature pyramid and b) can thus make predictions employing small head networks shared over all feature scales.
This leads to richly encoded scale information, strong occlusion handling and fast inference.

\begin{figure*}[thpb]
   \centering
   \includegraphics[width=2.0\columnwidth]{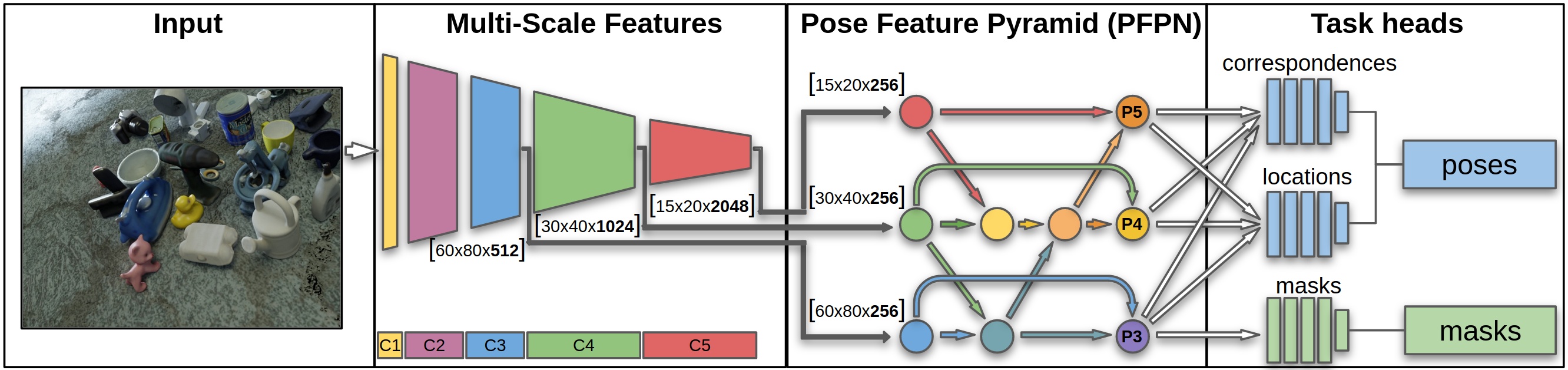}
   \caption{PyraPose network design: Multi-resolution feature aggregation from different backbone stages using a specialized Pose Feature Pyramid Network (PFPN) and task heads for patch-wise pose hypotheses creation and mask prediction.}
   \label{fig:arch}
\end{figure*}

\subsection{Feature Pyramids}

Hierarchical feature aggregation over various spatial resolutions provides local to global object appearance cues, thus producing a strong set of object hypotheses. 
In order to fully utilize this information, approaches such as Faster-RCNN~\cite{ren2015faster} use translation-invariant anchor boxes.
These reference locations and bounding box priors enable object hypotheses to be created independently and simultaneously from multiple partial views. 
Based on this anchor design,~\cite{liu2016ssd} extends the idea by aggregating feature maps of varying resolutions. For fusing features over different spatial resolutions, the Feature Pyramid Network (FPN)~\cite{lin2017feature} is currently the most used approach. 
By aggregating features using a top-down pathway then upsampling lower-resolved feature maps for feature fusion, FPN improves performance over using the feature maps from a single image resolution.
Building on FPN, the Path Aggregating Network~\cite{liu2018path} adds a bottom-up pathway. 
For each spatial resolution's proposal, features of all pyramid levels vote for predictions.
Recently,~\cite{ghiasi2019fpn} trained Neural Architecture Search~\cite{zoph2016neural} to find the optimal feature pyramid network design by learning the process given a search space.
In~\cite{tan2019efficientdet}, an efficient modification of FPN is presented in which the introduction of skip connections leads to a more lightweight and better performing variant.

These methods are effective for object detection and we translate the concept to pose estimation with our proposed PFPN.
By leveraging recent findings for feature pyramids and designing PFPN for pose estimation, we achieve accurate pose estimates.



\subsection{Synthetic-to-real Transfer}

The two most common methods for synthetic-to-real transfer for learning-based pose estimation are unsupervised domain adaptation and domain randomization~\cite{tobin2017domain}.
For unsupervised domain adaptation, Generative Adversarial Networks (GAN) are employed to map images from a source to a target domain~\cite{bousmalisSDEK16, Zakharov2018KeepIU}.
Bousmalis et al.~\cite{bousmalisSDEK16} train one model to map images from the synthetic source domain to the real target domain, followed by a second model to classify and estimate 3D poses.
In~\cite{Zakharov2018KeepIU}, a GAN is trained to transfer depth images from the augmented synthetic domain to the unaugmented synthetic domain for 3D rotation estimation. Thus, this method requires no real-world data during training time.
The authors of~\cite{rad18ACCV} learn a mapping between the synthetic and real-world features spaces for domain adaptation.
Recently, a self-supervised approach using differential rendering is proposed to overcome the requirement of annotated real-world images~\cite{wang2020self6d}.
However, while neither~\cite{rad18ACCV} nor~\cite{wang2020self6d} need annotated real-world images, both require a considerable amount of images in the target domain. 
Our approach works purely with synthetic data without the requirement of any real-world images during training. 
Domain adaptation is performed via randomizing the noise applied to the synthetic domain.

When randomizing domains, the image space is augmented by inducing noise~\cite{kehl2018, sundermeyer2018implicit, thalhammer} to create a synthetic domain that is a superset of the real-world domain. 
In~\cite{kehl2018}, rendered images are pasted onto images from MS COCO~\cite{lin2014microsoft} and trained for end-to-end object detection, classification and pose prediction using a network based on~\cite{liu2016ssd}.
In~\cite{zakharov2019dpod}, the authors follow the same strategy to estimate poses using an encoder-decoder architecture.
Sundermeyer et al.~\cite{sundermeyer2018implicit} train separate networks for detection and pose estimation.
For detection, the models from~\cite{liu2016ssd} and~\cite{linRetina} are used, then an encoder-decoder network is trained to reconstruct rendered images from augmented renderings. 
In~\cite{thalhammer}, synthetic depth images are augmented using a randomized camera model. Object detection, classification and correspondence prediction is done in a multi-task fashion.

In this work, we leverage recent RGB augmentation strategies to improve generalization.
We show that simple randomized augmentations, in combination with the proposed PFPN, are sufficient to translate learned features to the real world and to outperform the current state of the art.

\section{PyraPose}\label{sec:dmpose}


Our proposed approach for single-shot 6D object pose estimation is shown in Figure~\ref{fig:arch}. The first stage is an end-to-end trainable multi-task CNN. This network detects the object in the image and identifies its class.
Additionally, the 2D-3D correspondences between image space and the 3D camera coordinate system are regressed.
We further refer to these as \textit{correspondences}.
In the second stage, the Perspective-n-Point algorithm computes the 6D pose given the estimated correspondences.

We use only one shared network, independent of the amount of objects of interest. 
Therefore, only one model needs to be trained per dataset in contrast to the majority of state-of-the-art approaches~\cite{manhardt2019explaining, park2019pix2pose, peng2019pvnet, rad2017bb8, rad18ACCV, sundermeyer2018implicit, wang2019densefusion, zakharov2019dpod}.
This is advantageous, because the memory footprint, ease of application and time for training is little, and we only require one forward pass at runtime, independent of the application.
As a result, we achieve pose estimation at $\sim 26$ fps on an Nvidia Titan V.

The next section presents our proposed PFPN, followed by a presentation of the tasks for which we train our approach and and explanation how poses are estimated.

\subsection{Pose Feature Pyramid Network}\label{sec:PFPN}

The Pose Feature Pyramid Network is based on the principles presented in~\cite{ghiasi2019fpn, tan2019efficientdet}.
In Figure~\ref{fig:arch}, each node represents an add operation followed by a $3\times3$ convolution with stride $1$ and ReLU activation.
For object pose estimation, the variation of object scale is usually less than for object detection (e.g., MS COCO~\cite{lin2014microsoft}).
Furthermore, object pose estimation strongly relies on the features of early stages of the backbone.
Thus, our network design disregards higher levels of the feature pyramid, which are usually used for object detection (i.e., C6 and C7), and we only rely on the backbone levels \textit{C3}, \textit{C4}, and \textit{C5} of ResNet~\cite{he}. 
To support this claim, we show a comparison to FPN in Section~\ref{sec:experiments}.

In order to create pyramid level \textit{P3}, the low-level features from \textit{C3} and \textit{C4} are combined.
Following~\cite{ghiasi2019fpn}, all spatial resolutions from the backbone are used to create pyramid level \textit{P4}.
This aggregates features from each available pyramid level and applies a higher amount of convolutions than for \textit{P3} or \textit{P5}.
The smallest resolved backbone level, \textit{C5}, is minimally processed to keep the computational overhead low.
Cross connections are designed to provide information from all available backbone levels to create \textit{P5}.
Skip connections are used on \textit{P3} and \textit{P4} as is known to be beneficial~\cite{tan2019efficientdet}.

In general, we observe that combining only two inputs per add operation and applying a convolution to the output achieves best performance.
The aggregated features are then forwarded to the task heads.

\subsection{Task Heads}

\begin{figure}[t]
   \centering
   \includegraphics[width=0.8\columnwidth]{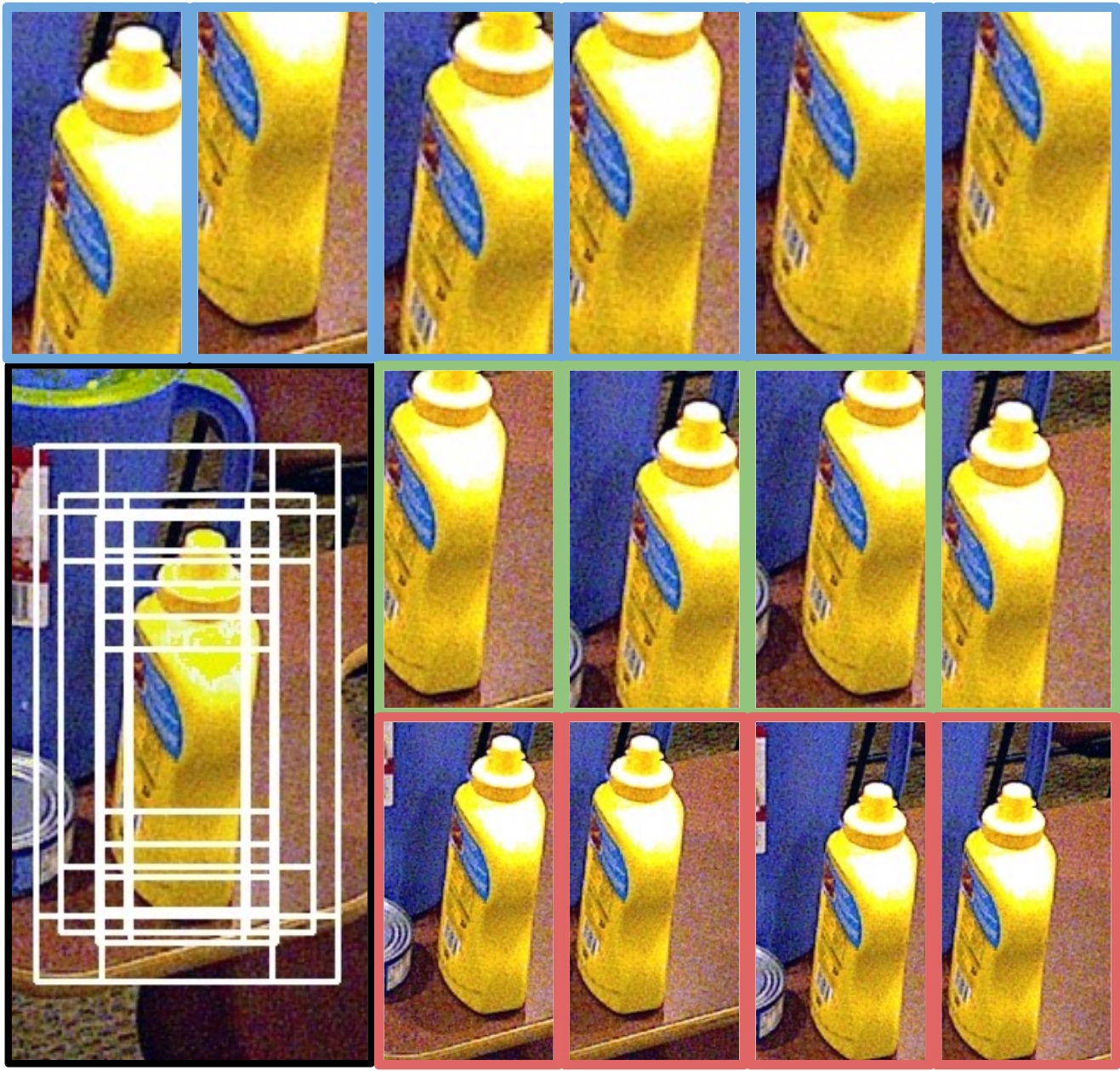}
   \caption{Image patches for pose hypotheses prediction.
   Lower left shows all the sampled priors (white boxes) used to predict the pose of YCB-video's~\cite{xiang2017posecnn} mustard bottle. Images on the right and top present some partial views in more detail.}
   \label{fig:priors}
\end{figure}

Our approach is supervised with 2D and 3D bounding boxes as well as segmentation masks.
The 3D bounding box and segmentation mask predictions update the trainable weights while the 2D bounding boxes are only used to compute image locations for pose hypotheses prediction.
The \textit{location} and \textit{correspondence} heads are used for pose estimation and the \textit{mask} head predicts the instance segmentation.
During training, 3D bounding boxes are projected into image space using the object pose in the camera frame and the camera intrinsics.
During inference, 3D bounding boxes are estimated in image space. Subsequently, PnP and the camera intrinsics are used to estimate the object's pose.

\subsubsection{Pose Hypotheses Prediction}
The \textit{location} and \textit{correspondence} heads classify image regions that contain an object and create 2D-3D correspondence hypotheses, respectively.
Figure~\ref{fig:priors} shows examples of such image regions.
The general procedure is similar to the concept of anchors for object detection~\cite{ren2015faster}.
For location classification, each pixel at every pyramid level acts as an anchor location and bounding box priors are created for each.
These priors are computed from a combination of different aspect ratios, scales and sizes dependant of the corresponding feature map resolution. 
The resulting priors that overlap the 2D bounding box with an Intersection-over-Union (IoU) of more than $0.5$ are considered as true locations.
Thus, in practice, ratios, scales and sizes are adjusted based on the expected bounding box sizes.
The \textit{correspondences} are normalized using the prior corner location as well as prior width and height.
Only true locations are used for updating the network's weight in the \textit{correspondence} heads.
All other anchors relate to the background class and thus do not contribute to \textit{correspondence} estimation.

Using anchor boxes for prior calculation forces the network to learn to produce hypotheses from different scales and partial views. This includes a small object to background ratio, which is shown to be beneficial in~\cite{oberweger2018making}.
The \textit{location} and \textit{correspondence} heads use all three feature map resolutions coming from PFPN for prediction making (i.e., \textit{P3}, \textit{P4}, \textit{P5}).

\subsubsection{Mask Prediction}

\begin{figure}[t]
   \centering
   \includegraphics[width=0.9\columnwidth]{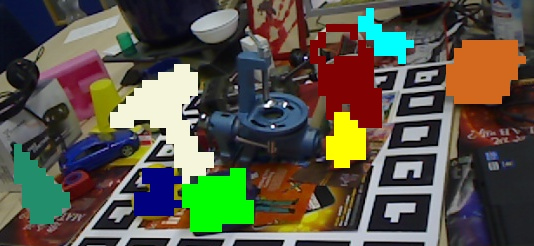}
   \caption{Coarse mask prediction on an image of Occlusion dataset~\cite{brachmann2014learning}. Predictions made with resolution 80$\times$60 pixels.}
   \label{fig:mask}
\end{figure}

Since depth data is widely available on robotic platforms, the Iterative Closest Point (ICP) algorithm can be used to refine initial pose estimates. 
Mask prediction is thus vital to achieve high quality performance with ICP.
We apply a similar approach to~\cite{hu2019segmentation} by predicting the mask at one eighth of the image resolution. This leads to usable masks while keeping the computational overhead low since no transposed convolutions are necessary to upsample the prediction. 
Conversely to the \textit{location} and the \textit{correspondence} heads, the \textit{mask} head takes only the features from \textit{P3} as input.
Figure~\ref{fig:mask} shows examples of predicted masks.

\subsubsection{Head Architectures and Losses}

All three task heads share the same architecture. 
Each consists of four convolution layers with a kernel size of $3\times3$, a stride of $1$ and ReLU activation, followed by another $3\times3$ convolution to regress the image locations of the eight 2D-3D correspondences with linear activation, or to classify using Sigmoid activation.
The first four convolutions in the \textit{location} and \textit{mask} heads extract 256 features per location. The same convolutions in the \textit{correspondence} head extract 512 features.
Additionally, the weights of the first four convolution in the \textit{correspondence} head are regularized using \textit{l2}-regularization with $\lambda$ set to 0.001.


For the \textit{correspondence} head, the orthogonality favoring loss of~\cite{thalhammer} is used with $\delta$ set to $0.8$. 
By punishing edge length differences of the ground truth and estimated 3D bounding boxes, the regression incorporates 3D information.
Both the \textit{location} and \textit{mask} heads are optimized using the focal loss~\cite{linRetina}, which is
commonly used to improve cross entropy loss when classes are imbalanced.
By down weighting examples that are easy to classify and focusing on difficult examples during loss calculation, this loss corrects misclassified examples. During training, the losses are normalized over batches, anchors (\textit{location} head) and anchor locations (\textit{mask} head).
Weights applied to the normalized loss of the \textit{correspondence}, \textit{location} and \textit{mask} heads are $0.125$, $1.0$ and $0.1$, respectively.

\subsection{Pose Estimation}

The set of predicted correspondence hypotheses are used for pose estimation.
We threshold the Sigmoid score outputs of the \textit{location} head to choose true anchor locations.
We use all anchor locations with a classification score above 0.5 for a specific object to generate the set \textit{correspondence} hypotheses. 
The dominant approach to recover the most likely reference frame transformation that aligns a set of 2D points to a set of noisy 3D points~\cite{hu2019segmentation, park2019pix2pose, peng2019pvnet, wang2019densefusion, zakharov2019dpod} is PnP.
We follow that trend and recover the 3D pose using the RANSAC version of PnP with 300 iterations.
When depth data is available, the inferred poses and masks are combined to refine the initial pose estimate with ICP.


\section{Experiments}\label{sec:experiments}

\begin{figure*}[h]
   \centering
      \begin{subfigure}
      \centering
      \includegraphics[width=0.6\columnwidth]{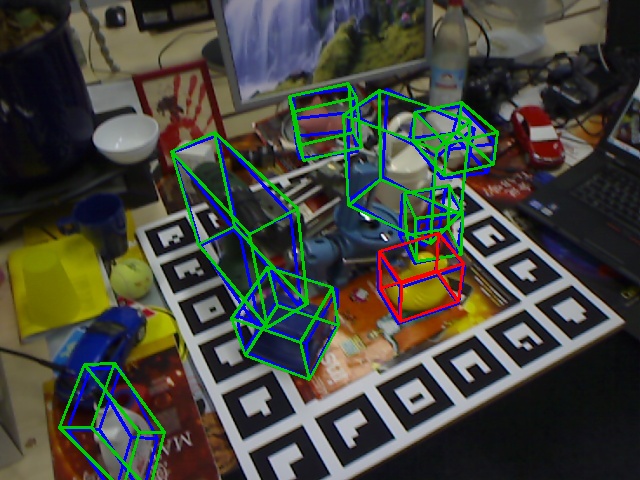}
   \end{subfigure}
   \begin{subfigure}
      \centering
      \includegraphics[width=0.6\columnwidth]{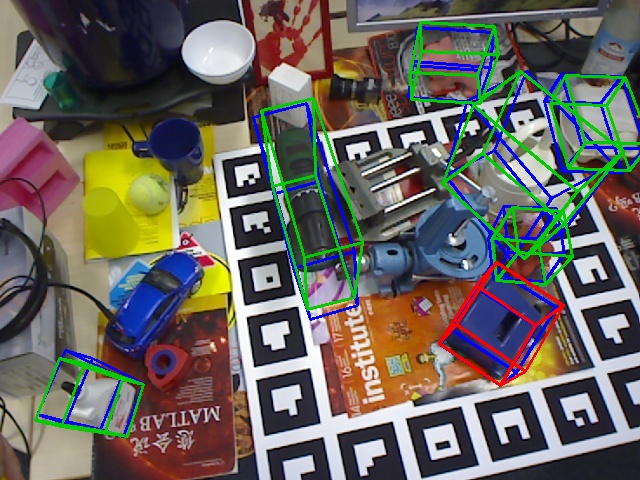}
   \end{subfigure}
   \begin{subfigure}
      \centering
      \includegraphics[width=0.6\columnwidth]{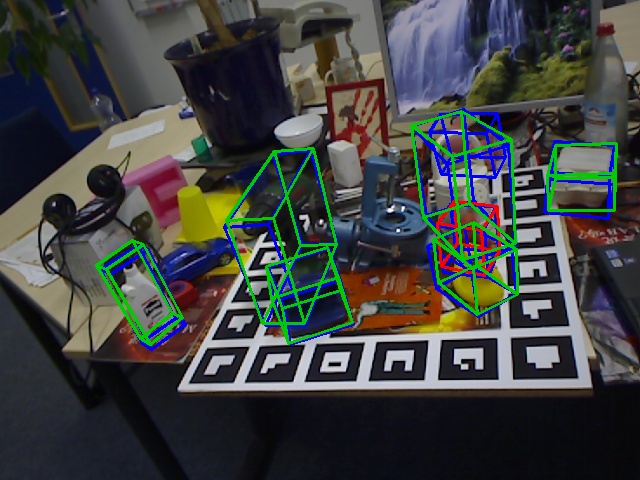}
   \end{subfigure}
   \caption{Examples of correctly estimated (green), incorrectly estimated (red) and ground truth (blue) poses on images from the Occlusion dataset~\cite{brachmann2014learning}.}
   \label{fig:viz}
\end{figure*}

This section presents experiments to evaluate our proposed method for pose estimation. We use three different datasets to compare to state of the art, show the advantage of using synthetic data to generalize to new domains, perform ablations of our architectural design and conduct grasping experiments to demonstrate the usability for real-world robotic applications.

\subsection{Datasets}

Single-object pose estimation of objects with little texture is provided with Linemod~\cite{hinterstoisser2012model},
which consists of approximately 1100 to 1300 images of cluttered scenes per object.
Linemod comes with 15 textured meshes of the objects of interest and corresponding annotations on the frame level.
We use the same subset of objects as~\cite{rad18ACCV, sundermeyer2018implicit, wang2020self6d, zakharov2019dpod}. 
Multi-object experiments for objects with little texture are performed on Occlusion~\cite{brachmann2014learning}.
This dataset is the test set of Linemod's Benchvise object featuring annotations for all dataset objects that appear in the images.


In addition to single- and multi-object experiments, we also present experiments on Homebrewed's~\cite{kaskman2019homebreweddb} second sequence to show that using synthetic data for training is better for generalizing to new settings.
The specific sequence from Homebrewed features three of Linemod's objects in a different setting, which includes scene setup, illumination and camera intrinsics. 
We show that PyraPose trained on synthetic images for the Linemod setting generalizes well to Homebrewed.

Lastly, in order to bridge from training on synthetic data to performing tasks in the physical world, we present grasping experiments on YCB-video objects~\cite{xiang2017posecnn} following the layout of the GRASPA 1.0 benchmark~\cite{Bottarel2020GRASPA1G}.

\subsection{Training Data and Image Augmentation}

Training images are rendered from complete scenes of textured 3D object models with realistic materials and lighting. The geometric configuration of objects and camera is generated using physics simulation. 
Synthesizing images is performed by physically-based rendering~\cite{hodan2019photorealistic}. 
For Linemod and Occlusion, we use the same training images.
The synthetic Linemod, Homebrewed and YCB-video datasets contain $50k$ samples taken from the BOP challenge~\cite{hodan2018bop}. 

In order to force domain generalization of our models, augmentations are applied during training.
Synthetic training images are randomly augmentations in random order to minimize the domain gap.
We apply randomized image augmentations similar to~\cite{sundermeyer2018implicit}.
However, we found slightly different ranges of the parameters to be beneficial.
We also vary the contrast, which is not a procedure performed in~\cite{sundermeyer2018implicit}.
Table~\ref{tab:augrgb} provides a list of applied augmentations and parameter value ranges.

\begin{table}[t!]
\caption{Training data augmentation parameters.}
\vspace{-1.0ex}
\label{tab:augrgb}
\begin{center}
\begin{tabular}{c|c|c}
\hline
Augmentation & Chance (per channel) & Range\\
\hline
gaussian blur & 0.2 & $\sigma \sim \mathcal{U}(0.0, 2.0)$\\
average/median/motion blur & 0.2 & $\sigma \sim \mathcal{U}(3, 7)$ \\
bilateral blur & 0.2 & $\sigma \sim \mathcal{U}(1, 7)$ \\
hue/saturation & 0.5 & $\mathcal{U}(-15, 15)$  \\
grayscale & 0.5 & $\mathcal{U}(0.0, 0.2)$  \\
add & 0.5 (0.5) & $\mathcal{U}(-0.04, 0.04)$  \\
multiply & 0.5 (0.5) & $\mathcal{U}(0.75, 1.25)$  \\
gamma contrast & 0.5 (0.5) &  $\mathcal{U}(0.75, 1.25)$ \\
sigmoid contrast & 0.5 (0.5) &  $\mathcal{U}(0, 10)$  \\
logarithmic contrast & 0.5 (0.5) & $\mathcal{U}(0.75, 1.0)$  \\
linear contrast & 0.5 (0.5) & $\mathcal{U}(0.7, 1.3)$  \\
\hline
\end{tabular}
\vspace{-2.0ex}
\end{center}
\end{table}

\subsection{Implementation Details}

Our networks are trained using the Adam optimizer with a learning rate initialized with $10^{-5}$ and a batch size of $8$.
The learning rate is decreased by one magnitude if the loss does not decrease for two epochs.
Results provided are after $200$ epochs of training on $50k$ synthetic RGB images on an Nvidia Geforce Titan V.
The weights of the first 14 convolutions are frozen similar to~\cite{manhardt2018deep, zakharov2019dpod}. 
Batch normalization layers are set to non-trainable and parameters $\gamma$ and $\beta$ are frozen.

\subsection{Evaluation Setting}

For object detection, we consider the existence of an object as true if the detection score of an anchor is above $0.5$.
For these true anchors, the estimated 2D-3D correspondences are considered as hypotheses.
For pose estimation, the ADD-recall is used~\cite{hinterstoisser2012model}.
This measures the average distance difference between the corresponding transformed object points from the estimated and the ground truth pose. 
For the objects \textit{Eggbox} and \textit{Glue} in Linemod, the symmetric version (ADDS) is used in which the closest point distance is used for calculation.
For all experiments, we report the percent of correctly estimated poses when the ADD-score is less than the standard threshold of $10 \%$ of the model diameter~\cite{hinterstoisser2012model}.

\subsection{Comparison to the State of the art}

We provide a quantitative comparison to recent learning-based object pose estimation methods of~\cite{sundermeyer2018implicit}, ~\cite{manhardt2019explaining} and~\cite{zakharov2019dpod} that use only synthetic data for training, and therefore, most similar to our approach, as well as against two winning methods of the 2019 BOP challenge~\cite{BOP19}, i.e.,~\cite{li2019CDPN} and~\cite{park2019pix2pose}. 

\subsubsection{Single-object scenario}

\begin{table}[t!]
\caption{ADD/S-recall on Linemod~\cite{hinterstoisser2012model} in comparison to synthetically trained methods. Last two columns show results of methods using test images without pose annotation for training. Objects with (*) are evaluated using ADDS.}
\label{tab:Sota}
\begin{center}

\begin{tabular}{|c|c|c|c|c||c|c|}
\hline
Training & \multicolumn{4}{|c||}{Synthetic} & \multicolumn{2}{|c|}{Real w/o pose}\\
\hline
Method & AAE & MHP & DPOD & PyraPose & DTPE & Self6D \\
& \cite{sundermeyer2018implicit} & \cite{manhardt2019explaining} & \cite{zakharov2019dpod} & (ours) & \cite{rad18ACCV} & \cite{wang2020self6d} \\
\hline
Ape & 4.2 & 11.9 & \textbf{35.1} & 22.8 & 19.8 & 38.9 \\
Benchv. & 22.9 & 66.2 & 59.4 & \textbf{78.6} & 69.0 & 75.2 \\
Cam & 32.9 & 22.4 & 15.5 & \textbf{56.5} & 37.6 & 36.9 \\
Can & 37.0 & 59.8 & 48.8 & \textbf{81.9} & 42.3 & 65.6 \\
Cat & 18.7 & 26.9 & 28.1 & \textbf{56.2} & 35.4 & 57.9 \\
Drill & 24.8 & 44.6 & 59.3 & \textbf{70.2} & 54.7 & 67.0 \\
Duck & 5.9 & 8.3 & 25.6 & \textbf{40.4} & 29.4 & 19.6 \\
Eggbox* & 81.0 & 55.7 & 51.2 & \textbf{84.4} & 85.2 & 99.0 \\
Glue* & 46.2 & 54.6 & 34.6 & \textbf{82.4} & 77.8 & 94.1 \\
Holep. & 18.2 & 15.5 & 17.7 & \textbf{42.6} & 36.0 & 16.2 \\
Iron & 35.1 & 60.8 & 84.7 & \textbf{86.4}  & 63.1 & 77.9 \\
Lamp & 61.2 & - & 45.0 & \textbf{62.0} & 75.1 & 68.2 \\
Phone & 36.3 & 34.4 & 20.9 & \textbf{59.5} & 44.8 & 50.1 \\
\hline
Avg. & 32.6 & 38.8 & 40.5 & \textbf{63.4} & 51.6 & 58.9 \\
\hline
\end{tabular}
\vspace{-2.0ex}
\end{center}
\end{table}

Table~\ref{tab:Sota} presents results on the Linemod dataset~\cite{hinterstoisser2012model}.
Our method significantly outperforms~\cite{manhardt2019explaining, sundermeyer2018implicit, zakharov2019dpod}\footnote{Results of~\cite{zakharov2019dpod} provided by the authors; these differ from their presented results because ADD-precision is presented while we present ADD-recall}, which are methods training on synthetic data with domain randomization applied for domain adaptation.
We the best results on average and for all individual objects except from the \textit{Ape}.
The worse results for the \textit{Ape} might be caused by the suboptimal anchor parameters for the small object scale.
The last two columns show results of~\cite{rad18ACCV} and~\cite{wang2020self6d}. Both methods use the test set of Linemod without pose annotation for unsupervised and self-supervised domain adaptation for training, respectively.
We achieve better performance in comparison even though no real images are used to train our network.

Our method provides strong estimates over all object scales due to the multi-scale nature when using feature pyramids.
Especially for robotic scenarios, our approach is beneficial because we train only one network for all objects, while all competing methods~\cite{manhardt2019explaining, rad18ACCV, sundermeyer2018implicit, wang2020self6d, zakharov2019dpod} train one network per object to gain better performance.
Our approach keeps memory load low, provides easier usage and still outperforms the state of the art.

\subsubsection{Multi-object scenario}

Table~\ref{tab:Occ} reports the ADD/S-recall for multi-object detection and pose estimation in comparison to the state-of-the-art learning-based pose estimation approaches of~\cite{li2019CDPN, park2019pix2pose, zakharov2019dpod}\footnote{Results are taken from~\cite{wang2020self6d}}.
These three methods employ encoder-decoder networks for pose hypotheses creation and we achieve significantly better performance using only synthetic data for training. 
Our multi-hypotheses creation scheme in combination with PFPN leads to robust prediction making under occlusion and effective multi-object pose estimation, as such achieving a relative improvement of $\sim$35~\% over~\cite{li2019CDPN}, which won the award for best RGB-only method in the 2019 BOP Challenge~\cite{BOP19}.

\begin{table}[t!]
\caption{ADD/S-recall on Occlusion~\cite{brachmann2014learning} in comparison to methods only using synthetic data for training. Objects with (*) are evaluated using ADDS.} 
\label{tab:Occ}
\begin{center}
\begin{tabular}{|c|c|c|c|c|}
\hline

Method & DPOD & CDPN & Pix2Pose & PyraPose \\ 
 & \cite{zakharov2019dpod} & \cite{li2019CDPN} & \cite{park2019pix2pose} & (ours) \\
\hline
Ape & 2.3 & \textbf{20.0} & 11.3 & 18.5 \\
Can & 4.0 & 15.1 & 18.5 & \textbf{46.4} \\ 
Cat & 1.2 & 16.4 & \textbf{17.1} & 11.7 \\ 
Drill & 10.5 & 5.0 & 34.5 & \textbf{48.2} \\ 
Duck & 7.2 & 22.2 & \textbf{25.3} & 19.4 \\ 
Eggbox* & 4.4 & \textbf{36.1} & 12.0 & 16.7 \\ 
Glue* & 12.9 & 27.9 & \textbf{30.8} & 30.7 \\ 
Holep. & 7.5 & 24.0 & 12.2 & \textbf{33.0} \\ 
\hline
Avg. & 6.3 & 20.8 & 20.2 & \textbf{28.1} \\ 
\hline
\end{tabular}
\end{center}
\end{table}

\subsubsection{Novel domain}


\begin{table}[ht]
\caption{ADD-recall on second test sequence of Homebrewed~\cite{kaskman2019homebreweddb} in comparison to the baseline method of~\cite{zakharov2019dpod}.}
\label{tab:HBDB}
\begin{center}
\begin{tabular}{|c|c|c|c|c||c|}
\hline

Method & Mesh origin & Benchv. & Drill & Phone & Avg. \\
\hline
DPOD \cite{zakharov2019dpod} & Hb \cite{kaskman2019homebreweddb} & 52.9 & 37.8 & 7.3 & 32.7 \\
PyraPose (ours) & Hb \cite{kaskman2019homebreweddb} & \textbf{62.9} & 22.6 & 38.5 & \textbf{41.3} \\
\hline
PyraPose (ours) & Lm \cite{hinterstoisser2012model} & 10.9 & \textbf{60.0} & \textbf{44.4} & 38.4 \\
\hline
\end{tabular}
\end{center}
\end{table}


\begin{figure}[ht]
   \centering
   \begin{subfigure}
      \centering
      \includegraphics[width=0.48\columnwidth]{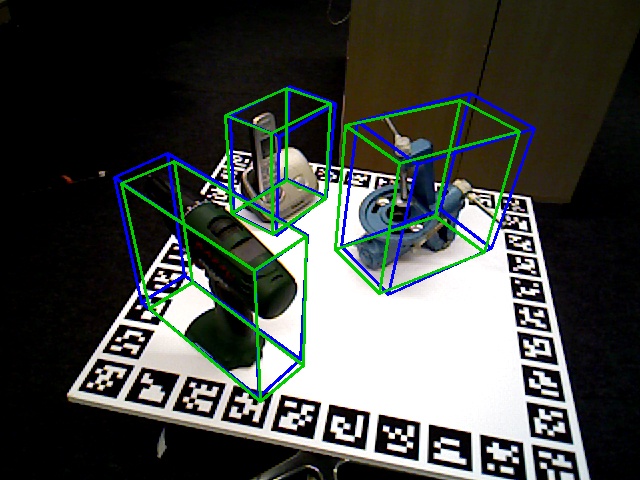}
   \end{subfigure}
   \begin{subfigure}
      \centering
      \includegraphics[width=0.48\columnwidth]{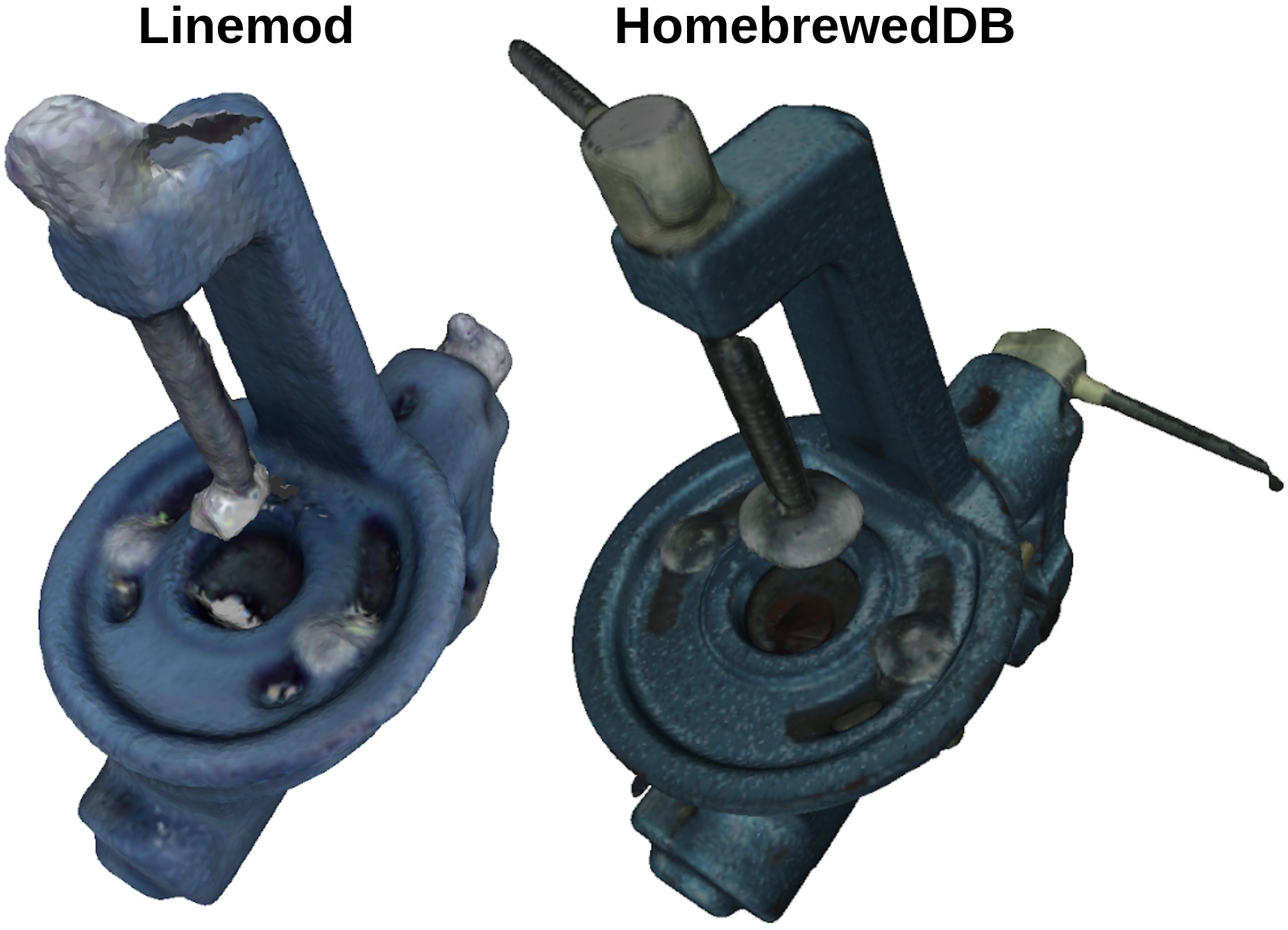}
   \end{subfigure}
   \caption{Left image shows pose estimates from HomebrewedDB~\cite{kaskman2019homebreweddb}. Ground truth in blue and correctly estimates poses in green. Right image shows the mesh of the~\textit{Benchvise} from~\cite{hinterstoisser2012model} and~\cite{kaskman2019homebreweddb}, respectively.}
   \label{fig:hb_pose}
\end{figure}


Pose estimators trained on real-world data tend to overfit to certain characteristics of the data that they are trained on~\cite{kaskman2019homebreweddb}. 
Training on synthetic data provides the benefit of translating better to novel domains, i.e., new places of deployment.
As such, we show results on the second sequence of Homebrewed~\cite{kaskman2019homebreweddb}, which has images with pose annotations of Linemod's~\textit{Benchvise},~\textit{Drill} and~\textit{Phone}.
DPOD~\cite{zakharov2019dpod}\footnote{Results are taken from~\cite{wang2020self6d}} is used by the authors of Homebrewed~\cite{kaskman2019homebreweddb} as a baseline for the dataset. This method is trained on synthetic images using the provided meshes.
The results provided in Table~\ref{tab:HBDB}, using the ADD-recall on the validation set of~\cite{hodan2018bop}, show that PyraPose outperforms DPOD when trained with synthetic data rendered for the Homebrewed setting (using meshes of all dataset objects for training). Figure~\ref{fig:hb_pose} (left) shows an example of pose estimates.
 
The last row in Table~\ref{tab:HBDB} gives results of our method when reusing our network trained for Linemod~\cite{hinterstoisser2012model}. 
In order to use our model trained on Linemod, the corresponding object models are manually aligned to compute the ADD-score, since the coordinate frame origin of all three models differ in~\cite{hinterstoisser2012model} and~\cite{kaskman2019homebreweddb}. 
The resulting relative transformations between models are added as a constant offset to Homebrewed's test annotations.
Our method effectively transitions to the novel place of deployment, i.e., overcomes the domain shift by training on synthetic data, which is highly useful in robotics.
The results for the \textit{Benchvise} are underwhelming, however, and we conjecture that the reason is due to the significantly worse reconstruction used to render the data when trained for Linemod, as shown in Figure~\ref{fig:hb_pose} (right).

\subsubsection{Runtime}

A comparison of runtime is provided against other single-shot object pose estimators that are designed for fast inference.
A forward pass of our method takes on average 39 ms on an NVIDIA Titan V ($\sim$26 fps), excluding PnP. RANSAC-PnP produces an overhead of approx 1 ms per detected object.
In comparison, AAE~\cite{sundermeyer2018implicit} computes estimates at a rate of 13 fps when using RetinaNet~\cite{linRetina} as backbone (used for the results in Table~\ref{tab:Sota}) and DPOD at 33 fps. 
The best performing methods~\cite{park2019pix2pose} and~\cite{li2019CDPN} are significantly slower, computing estimates at a rate of $\sim$0.8 and $\sim$1 fps, respectively, without refinement\footnote{Inference times taken from~\cite{BOP19}}.

\subsection{Ablation Study}

We perform an ablation study by comparing our pose estimation pipeline using the the proposed PFPN, FPN~\cite{lin2017feature} and using no feature aggregation (None).
When using no feature aggregation, feature maps from the backbone stages are each projected to 256 features by separate $1\times1$ convolutions with our task heads for estimation.

Table~\ref{tab:FPN_OC} shows that our feature aggregation scheme, that focuses on low-level features, improves pose estimation performance compared to FPN.
For small objects, the location and correspondence estimation are primarily computed from feature map resolution \textit{P3}, i.e., from low-level features, which results in similar performance. However, for large objects, the feature map resolution \textit{P5} is primarily used for estimation and PFPN performs significantly better, which demonstrates the advantage of focusing on low-level features. Using no feature aggregation has the worst performance and supports the hypothesis that feature aggregation is a useful tool for pose estimation.

\begin{table}[t!]
\caption{Multi-scale feature aggregation comparison on Linemod~\cite{hinterstoisser2012model} and Occlusion~\cite{brachmann2014learning}}
\label{tab:FPN_OC}
\begin{center}
\begin{tabular}{|c|c|c|c|c|c|c|}
\hline
Feature & \multicolumn{3}{|c|}{Linemod \cite{hinterstoisser2012model}} & \multicolumn{3}{|c|}{Occlusion \cite{brachmann2014learning}}\\
aggregation & None & FPN & PFPN & None & FPN & PFPN \\
 &  & \cite{lin2017feature} & (ours) &  & \cite{lin2017feature} & (ours) \\
\hline
Ape & 15.9 & \textbf{28.6}  & 22.8 & 14.1  & \textbf{19.9} & 18.5 \\
Benchv. & 70.7 & 73.1 & \textbf{78.6} & -  & - & - \\
Cam & 27.0 & 55.4 & \textbf{56.5} & -  & - & - \\
Can & 41.6 & 73.1  & \textbf{81.9} & 29.7 & 37.7 & \textbf{46.4} \\
Cat & 45.3 & 52.2 & \textbf{56.2} & 13.2 & \textbf{11.7} & \textbf{11.7} \\
Drill & 47.8 & 63.6 & \textbf{70.2} & 31.1 & 46.1 & \textbf{48.2} \\
Duck & 33.5 & 32.9 & \textbf{40.4} & 11.2 & \textbf{20.5} & 19.4 \\
Eggbox & 60.7 & 81.2 & \textbf{84.4} & 14.1 & 16.3 & \textbf{16.7} \\
Glue & 43.6 & \textbf{91.5} & 82.4 & 24.0 & \textbf{31.1} & 30.7 \\
Holep. & 33.8 & 39.5 & \textbf{42.6} & 33.1 & \textbf{34.6} & 33.0\\
Iron & 79.9 & 83.2 & \textbf{86.4} & - & - & - \\
Lamp & 66.1 & 57.2 & \textbf{62.0} & - & - & - \\
Phone & 39.2 & 51.9 & \textbf{59.5} & - & - & - \\
\hline
Avg. & 46.5 & 60.2 & \textbf{63.4} & 21.3 & 27.3 & \textbf{28.1} \\
\hline
\end{tabular}
\end{center}
\end{table}


\begin{table}[ht]
\caption{Object grasping of YCB-video~\cite{xiang2017posecnn} objects. 10 trials are performed per object. Comparison is given for grasping without refinement (w/o ICP) and when refining the initial pose with instance segmentation and ICP (/w ICP).}
\label{tab:GraspA}
\begin{center}
\begin{tabular}{|c|c|c|c|c|c||c|}
\hline
Object & Mustard & Jell-o & Spam & Banana & Foam & Success \\
\hline
w/o ICP & 10 & 1 & 9 & 7 & 3 & 60\% \\
/w ICP & 10 & 5 & 10 & 3 & 7 & 70\% \\
\hline
\end{tabular}
\end{center}
\end{table}


\begin{figure}[ht]
   \centering
   \vspace{-1ex}
    \includegraphics[width=0.94\columnwidth]{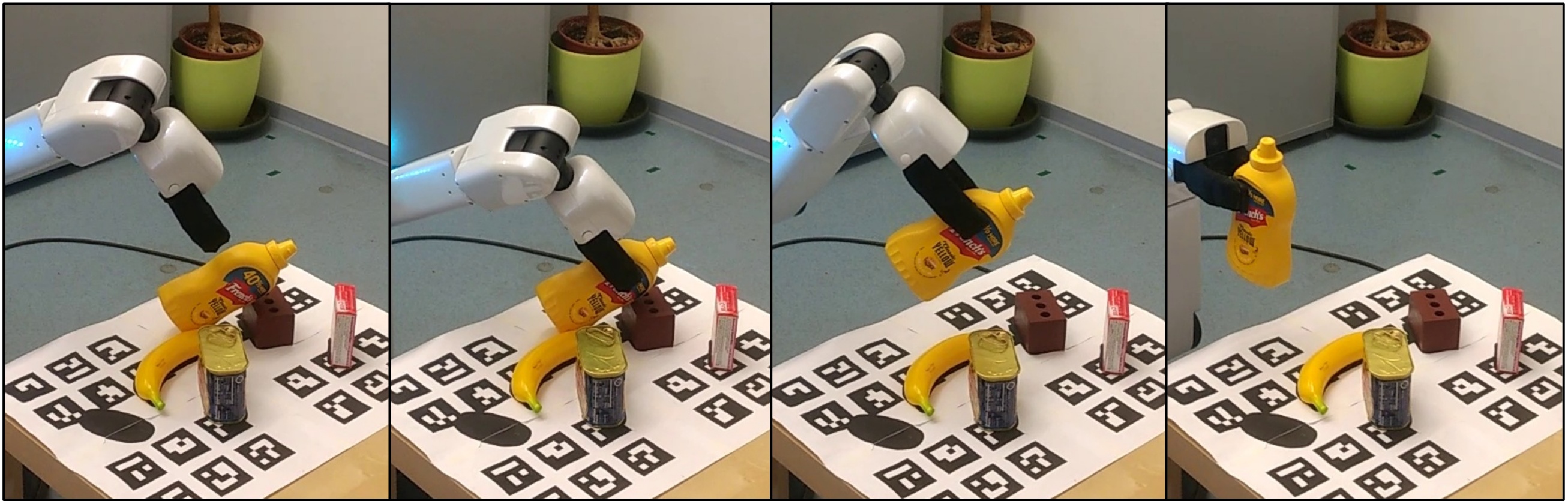}
   \vspace{-1ex}
   \caption{Mustard bottle is successfully grasped from a pose supported by foam and remains stable in the gripper.}
   \label{fig:grasp_ycbv}
\end{figure}


\subsection{Grasping in the Real World}

Grasping experiments are conducted with the Toyota HSR~\cite{Yamamoto2018hsr, Yamamoto2019hsr} to demonstrate the ability to transition to real-world applications using only synthetic data for training.
Experiments use the printable GRASPA benchmark layout~0~\cite{Bottarel2020GRASPA1G}, which consists of five objects from YCB-video~\cite{xiang2017posecnn}: mustard bottle (mustard), gelatine box (jell-o), potted meat can (spam), banana and foam brick (foam).
The grasping pipeline is the same as in~\cite{VeRefineBauer20}. Each object has annotated grasp poses that are transformed to the robot base frame using the estimated object pose.
Based on the potential grasp poses, multiple trajectories are calculated and the first collision-free trajectory found is executed. A grasp is successful if the object is lifted and remains stable in the gripper.


Table~\ref{tab:GraspA} compares grasping performance using only the initial pose estimate and refining that estimate using the instance segmentation and ICP. Figure~\ref{fig:grasp_ycbv} shows an example of a successful grasp of mustard. Good performance is achieved for many objects, in particular, mustard, spam and foam (after refinement).
The poor performance for jell-o is explained by the significantly different texture of the real object compared to the model used to render the training data.
Interestingly, the grasp success for the banana drops when using refinement. This is explained by the observation that the locally optimal output of ICP often resulted in a pose that was rotated towards the table, which led to many grasp trajectories protruding the table plane. For the other objects, however, the segmentation followed by ICP improved performance, which shows the advantage of computing the masks.


\section{Conclusion}\label{sec:conclusion}


This paper presented a deep architecture for object pose estimation under domain shift.
Our proposed approach extracts and aggregates multi-resolution features for object pose estimation with a novel feature pyramid network for better synthetic-to-real transfer and occlusion handling.
Experiments show that our method outperforms the state of the art on multiple datasets even though only using one shared network, independent of the number of objects of interest. Furthermore, grasping experiments in the real world demonstrate the fitness of synthetic data for generalizing to novel environments.

Future work should address the performance drop that occurs due to imperfect object meshes since their quality is a major limitation of using synthetic data for training.
In order to even further improve performance using RGB only, a current research trend is to refine initial pose estimates without using depth data and ICP. Efforts should be made to transition the idea of coalescing multi-scale features for pose refinement.










\bibliographystyle{IEEEtran}
\bibliography{bibli}

\begin{thebibliography}{10}
\providecommand{\url}[1]{#1}
\csname url@samestyle\endcsname
\providecommand{\newblock}{\relax}
\providecommand{\bibinfo}[2]{#2}
\providecommand{\BIBentrySTDinterwordspacing}{\spaceskip=0pt\relax}
\providecommand{\BIBentryALTinterwordstretchfactor}{4}
\providecommand{\BIBentryALTinterwordspacing}{\spaceskip=\fontdimen2\font plus
\BIBentryALTinterwordstretchfactor\fontdimen3\font minus
  \fontdimen4\font\relax}
\providecommand{\BIBforeignlanguage}[2]{{%
\expandafter\ifx\csname l@#1\endcsname\relax
\typeout{** WARNING: IEEEtran.bst: No hyphenation pattern has been}%
\typeout{** loaded for the language `#1'. Using the pattern for}%
\typeout{** the default language instead.}%
\else
\language=\csname l@#1\endcsname
\fi
#2}}
\providecommand{\BIBdecl}{\relax}
\BIBdecl

\bibitem{kaskman2019homebreweddb}
R.~Kaskman, S.~Zakharov, I.~Shugurov, and S.~Ilic, ``{HomebrewedDB}: {RGB-D}
  dataset for {6D} pose estimation of {3D} objects,'' in \emph{Proceedings of
  the IEEE International Conference on Computer Vision Workshops}, 2019.

\bibitem{hu2019segmentation}
Y.~Hu, J.~Hugonot, P.~Fua, and M.~Salzmann, ``Segmentation-driven {6D} object
  pose estimation,'' in \emph{Proceedings of the IEEE Conference on Computer
  Vision and Pattern Recognition}, 2019, pp. 3385--3394.

\bibitem{li2019CDPN}
Z.~{Li}, G.~{Wang}, and X.~{Ji}, ``{CDPN}: {C}oordinates-based disentangled
  pose network for real-time {RGB}-based {6-DoF} object pose estimation,'' in
  \emph{Proceedings of the IEEE/CVF International Conference on Computer
  Vision}, 2019, pp. 7677--7686.

\bibitem{park2019pix2pose}
K.~Park, T.~Patten, and M.~Vincze, ``{Pix2Pose}: {P}ixel-wise coordinate
  regression of objects for {6D} pose estimation,'' in \emph{Proceedings of the
  IEEE International Conference on Computer Vision}, 2019, pp. 7668--7677.

\bibitem{peng2019pvnet}
S.~Peng, Y.~Liu, Q.~Huang, X.~Zhou, and H.~Bao, ``{PVNet}: {P}ixel-wise voting
  network for {6DoF} pose estimation,'' in \emph{Proceedings of the IEEE
  Conference on Computer Vision and Pattern Recognition}, 2019, pp. 4561--4570.

\bibitem{zakharov2019dpod}
S.~Zakharov, I.~Shugurov, and S.~Ilic, ``{DPOD}: {6D} pose object detector and
  refiner,'' in \emph{Proceedings of the IEEE International Conference on
  Computer Vision}, 2019, pp. 1941--1950.

\bibitem{liu2016ssd}
W.~Liu, D.~Anguelov, D.~Erhan, C.~Szegedy, S.~Reed, C.-Y. Fu, and A.~C. Berg,
  ``Ssd: Single shot multibox detector,'' in \emph{Proceedings of the European
  Conference on Computer Vision}, 2016, pp. 21--37.

\bibitem{ghiasi2019fpn}
G.~Ghiasi, T.-Y. Lin, and Q.~V. Le, ``{NAS-FPN}: {L}earning scalable feature
  pyramid architecture for object detection,'' in \emph{Proceedings of the IEEE
  Conference on Computer Vision and Pattern Recognition}, 2019, pp. 7036--7045.

\bibitem{lin2017feature}
T.-Y. Lin, P.~Doll{\'a}r, R.~Girshick, K.~He, B.~Hariharan, and S.~Belongie,
  ``Feature pyramid networks for object detection,'' in \emph{Proceedings of
  the IEEE Conference on Computer Vision and Pattern Recognition}, 2017, pp.
  2117--2125.

\bibitem{linRetina}
T.-Y. Lin, P.~Goyal, R.~B. Girshick, K.~He, and P.~Doll{\'a}r, ``Focal loss for
  dense object detection,'' in \emph{Proceedings of IEEE International
  Conference on Computer Vision}, 2017, pp. 2999--3007.

\bibitem{liu2018path}
S.~Liu, L.~Qi, H.~Qin, J.~Shi, and J.~Jia, ``Path aggregation network for
  instance segmentation,'' in \emph{Proceedings of the IEEE Conference on
  Computer Vision and Pattern Recognition}, 2018, pp. 8759--8768.

\bibitem{tan2019efficientdet}
M.~Tan, R.~Pang, and Q.~V. Le, ``{EfficientDet}: {S}calable and efficient
  object detection,'' in \emph{Proceedings of the IEEE/CVF Conference on
  Computer Vision and Pattern Recognition}, 2020, pp. 10\,778--10\,787.

\bibitem{he2019pvn3d}
Y.~He, W.~Sun, H.~Huang, J.~Liu, H.~Fan, and J.~Sun, ``{PVN3D}: {A} deep
  point-wise {3D} keypoints voting network for {6DoF} pose estimation,'' in
  \emph{Proceedings of the IEEE/CVF Conference on Computer Vision and Pattern
  Recognition}, 2020, pp. 11\,629--11\,638.

\bibitem{Hodan2020EPOSE6}
T.~Hodan, D.~Bar{\'a}th, and J.~Matas, ``{EPOS}: {E}stimating {6D} pose of
  objects with symmetries,'' \emph{Proceedings of the IEEE/CVF Conference on
  Computer Vision and Pattern Recognition}, pp. 11\,700--11\,709, 2020.

\bibitem{oberweger2018making}
M.~Oberweger, M.~Rad, and V.~Lepetit, ``Making deep heatmaps robust to partial
  occlusions for {3D} object pose estimation,'' in \emph{Proceedings of the
  European Conference on Computer Vision}, 2018, pp. 119--134.

\bibitem{wang2019densefusion}
C.~Wang, D.~Xu, Y.~Zhu, R.~Mart{\'\i}n-Mart{\'\i}n, C.~Lu, L.~Fei-Fei, and
  S.~Savarese, ``{DenseFusion}: {6D} object pose estimation by iterative dense
  fusion,'' in \emph{Proceedings of the IEEE Conference on Computer Vision and
  Pattern Recognition}, 2019, pp. 3343--3352.

\bibitem{rad2017bb8}
M.~Rad and V.~Lepetit, ``{BB8}: {A} scalable, accurate, robust to partial
  occlusion method for predicting the {3D} poses of challenging objects without
  using depth,'' in \emph{Proceedings of the IEEE International Conference on
  Computer Vision}, 2017, pp. 3828--3836.

\bibitem{tekin2018real}
B.~Tekin, S.~N. Sinha, and P.~Fua, ``Real-time seamless single shot {6D} object
  pose prediction,'' in \emph{Proceedings of the IEEE Conference on Computer
  Vision and Pattern Recognition}, 2018, pp. 292--301.

\bibitem{ren2015faster}
S.~Ren, K.~He, R.~Girshick, and J.~Sun, ``Faster {R-CNN}: {T}owards real-time
  object detection with region proposal networks,'' in \emph{Advances in Neural
  Information Processing Systems}, 2015, pp. 91--99.

\bibitem{zoph2016neural}
B.~Zoph and Q.~V. Le, ``Neural architecture search with reinforcement
  learning,'' \emph{International Conference on Learning Representations},
  2016.

\bibitem{tobin2017domain}
J.~Tobin, R.~Fong, A.~Ray, J.~Schneider, W.~Zaremba, and P.~Abbeel, ``Domain
  randomization for transferring deep neural networks from simulation to the
  real world,'' in \emph{Proceedings of the IEEE/RSJ International Conference
  on Intelligent Robots and Systems}, 2017, pp. 23--30.

\bibitem{bousmalisSDEK16}
K.~Bousmalis, N.~Silberman, D.~Dohan, D.~Erhan, and D.~Krishnan, ``Unsupervised
  pixel-level domain adaptation with generative adversarial networks,''
  \emph{Proceedings of the IEEE Conference on Computer Vision and Pattern
  Recognition}, pp. 95--104, 2016.

\bibitem{Zakharov2018KeepIU}
S.~Zakharov, B.~Planche, Z.~Wu, A.~Hutter, H.~Kosch, and S.~Ilic, ``Keep it
  unreal: {B}ridging the realism gap for {2.5D} recognition with geometry
  priors only,'' \emph{Proceedings of the International Conference on 3D
  Vision}, pp. 1--11, 2018.

\bibitem{rad18ACCV}
M.~Rad, M.~Oberweger, and V.~Lepetit, ``Domain transfer for {3D} pose
  estimation from color images without manual annotations,'' in
  \emph{Proceedings of the Asian Conference on Computer Vision}, 2018.

\bibitem{wang2020self6d}
G.~Wang, F.~Manhardt, J.~Shao, X.~Ji, N.~Navab, and F.~Tombari, ``{Self6D}:
  {S}elf-supervised monocular {6D} object pose estimation,'' in
  \emph{Proceedings of the European Conference on Computer Vision}, 2020.

\bibitem{kehl2018}
W.~Kehl, F.~Manhardt, F.~Tombari, S.~Ilic, and N.~Navab, ``{SSD-6D}: {M}aking
  {RGB}-based {3D} detection and {6D} pose estimation great again,''
  \emph{Proceedings of the IEEE International Conference on Computer Vision},
  pp. 1530--1538, 2017.

\bibitem{sundermeyer2018implicit}
M.~Sundermeyer, Z.-C. Marton, M.~Durner, M.~Brucker, and R.~Triebel, ``Implicit
  {3D} orientation learning for {6D} object detection from {RGB} images,'' in
  \emph{Proceedings of the European Conference on Computer Vision}, 2018, pp.
  699--715.

\bibitem{thalhammer}
S.~Thalhammer, T.~Patten, and M.~Vincze, ``{SyDPose}: {O}bject detection and
  pose estimation in cluttered real-world depth images trained using only
  synthetic data,'' in \emph{Proceedings of the International Conference on 3D
  Vision}, 2019, pp. 106--115.

\bibitem{lin2014microsoft}
T.-Y. Lin, M.~Maire, S.~Belongie, J.~Hays, P.~Perona, D.~Ramanan,
  P.~Doll{\'a}r, and C.~L. Zitnick, ``{Microsoft COCO}: {C}ommon objects in
  context,'' in \emph{Proceedings of the European Conference on Computer
  Vision}, 2014, pp. 740--755.

\bibitem{manhardt2019explaining}
F.~Manhardt, D.~M. Arroyo, C.~Rupprecht, B.~Busam, T.~Birdal, N.~Navab, and
  F.~Tombari, ``Explaining the ambiguity of object detection and {6D} pose from
  visual data,'' in \emph{Proceedings of the IEEE International Conference on
  Computer Vision}, 2019, pp. 6841--6850.

\bibitem{he}
K.~He, X.~Zhang, S.~Ren, and J.~Sun, ``Deep residual learning for image
  recognition,'' in \emph{Proceedings of IEEE Conference on Computer Vision and
  Pattern Recognition}, 2016, pp. 770--778.

\bibitem{xiang2017posecnn}
Y.~Xiang, T.~Schmidt, V.~Narayanan, and D.~Fox, ``{PoseCNN}: {A} convolutional
  neural network for {6D} object pose estimation in cluttered scenes,'' in
  \emph{Proceedings of Robotics: Science and Systems}, 2018.

\bibitem{brachmann2014learning}
E.~Brachmann, A.~Krull, F.~Michel, S.~Gumhold, J.~Shotton, and C.~Rother,
  ``Learning {6D} object pose estimation using {3D} object coordinates,'' in
  \emph{Proceedings of the European Conference on Computer Vision}, 2014, pp.
  536--551.

\bibitem{hinterstoisser2012model}
S.~Hinterstoisser, V.~Lepetit, S.~Ilic, S.~Holzer, G.~Bradski, K.~Konolige, and
  N.~Navab, ``Model based training, detection and pose estimation of
  texture-less {3D} objects in heavily cluttered scenes,'' in \emph{Proceedings
  of the Asian Conference on Computer Vision}, 2012, pp. 548--562.

\bibitem{Bottarel2020GRASPA1G}
F.~Bottarel, G.~Vezzani, U.~Pattacini, and L.~Natale, ``{GRASPA} 1.0: {GRASPA}
  is a robot arm grasping performance benchmark,'' \emph{IEEE Robotics and
  Automation Letters}, vol.~5, no.~2, pp. 836--843, 2020.

\bibitem{hodan2019photorealistic}
T.~Hoda{\v{n}}, V.~Vineet, R.~Gal, E.~Shalev, J.~Hanzelka, T.~Connell,
  P.~Urbina, S.~Sinha, and B.~Guenter, ``Photorealistic image synthesis for
  object instance detection,'' \emph{Proceedings of the IEEE International
  Conference on Image Processing}, 2019.

\bibitem{hodan2018bop}
T.~Hoda{\v{n}}, M.~Sundermeyer, B.~Drost, Y.~Labb{\'e}, E.~Brachmann,
  F.~Michel, C.~Rother, and J.~Matas, ``{BOP} challenge 2020 on {6D} object
  localization,'' \emph{Proceedings of the European Conference on Computer
  Vision Workshops}, 2020.

\bibitem{manhardt2018deep}
F.~Manhardt, W.~Kehl, N.~Navab, and F.~Tombari, ``Deep model-based {6D} pose
  refinement in {RGB},'' in \emph{Proceedings of the European Conference on
  Computer Vision}, 2018, pp. 800--815.

\bibitem{BOP19}
``{BOP}: Benchmark for 6d object pose estimation,''
  \url{https://bop.felk.cvut.cz/home/}, accessed: 2020-10-14.

\bibitem{Yamamoto2018hsr}
T.~{Yamamoto}, K.~{Terada}, A.~{Ochiai}, F.~{Saito}, Y.~{Asahara}, and
  K.~{Murase}, ``Development of the research platform of a domestic mobile
  manipulator utilized for international competition and field test,'' in
  \emph{Proceedings of the IEEE/RSJ International Conference on Intelligent
  Robots and Systems}, 2018, pp. 7675--7682.

\bibitem{Yamamoto2019hsr}
T.~Yamamoto, K.~Terada, A.~Ochiai, F.~Saito, Y.~Asahara, and K.~Murase,
  ``Development of human support robot as the research platform of a domestic
  mobile manipulator,'' \emph{ROBOMECH Journal}, vol.~6, no.~4, pp. 1--15,
  2019.

\bibitem{VeRefineBauer20}
D.~Bauer, T.~Patten, and M.~Vincze, ``{VeREFINE}: {I}ntegrating object pose
  verification with physics-guided iterative refinement,'' \emph{{IEEE}
  Robotics Automation Letters}, vol.~5, no.~3, pp. 4289--4296, 2020.

\end{thebibliography}

\end{document}